\documentclass[runningheads]{llncs}
\usepackage[T1]{fontenc}
\usepackage{graphicx}
\usepackage{booktabs, multirow, amsmath}
\usepackage{adjustbox}
\usepackage{bm}
\usepackage{fix-cm}
\usepackage{pgfplots}
\pgfplotsset{compat=1.18}
\usepackage{amssymb}
\usepackage{xcolor}
\usepackage{tikz}
\usepackage{orcidlink}

\usetikzlibrary{positioning, arrows.meta, fit, backgrounds,calc}
\definecolor{clrGrayFill}{RGB}{241,239,232}
\definecolor{clrGrayStroke}{RGB}{180,178,169}
\definecolor{clrPurpleFill}{RGB}{238,237,254}
\definecolor{clrPurpleStroke}{RGB}{175,169,236}
\definecolor{clrPurpleText}{RGB}{38,33,92}
\definecolor{clrTealFill}{RGB}{225,245,238}
\definecolor{clrTealStroke}{RGB}{93,202,165}
\definecolor{clrTealText}{RGB}{4,52,44}
\definecolor{clrAmberFill}{RGB}{250,238,218}
\definecolor{clrAmberStroke}{RGB}{186,117,23}
\definecolor{clrAmberText}{RGB}{65,36,2}
\definecolor{clrBlueFill}{RGB}{230,241,251}
\definecolor{clrBlueStroke}{RGB}{24,95,165}
\definecolor{clrBlueText}{RGB}{4,44,83}

\begin{document}

\title{Post-Operative Glioma Segmentation via Loss Stabilization, Normalization and Subspace Attention}
\titlerunning{Post-Operative Glioma Segmentation}
\author{Alexandru Crișan\inst{1}\orcidlink{0009-0001-6246-8886} \and
Diana Borza\inst{1}}
\authorrunning{A. Crișan and D. Borza}
\institute{Babeș-Bolyai University, Cluj-Napoca, Romania \\
\email{alexandru.crisan3@stud.ubbcluj.ro, diana.borza@ubbcluj.ro}}

\maketitle

\begin{abstract}
Tracking residual tumor after surgery is essential for catching recurrence early, but automating post-operative glioma segmentation remains a difficult task.  Although transformer-based architectures, such as 
SwinUNETR, achieved impressive results, few studies test how well they generalize across clinical protocols. In this paper, we conduct an ablation study on the MU-GLIOMA-POST and UCSF-ALPTDG datasets and show that the standard Generalized Dice Loss (GDL) is unstable under domain shift: the Whole Lesion (WL) Dice drops from $0.88$ on the internal validation set to $0.73$ on the external UCSF test set. To address this, we pair brain-masked percentile normalization with voxel-level contrastive learning.
We also propose a Subspace-Aware Class Attention (SACA) module that re-calibrates the bottleneck features and raises Enhancing Tumor (ET) sensitivity by 8\% (9.1\% relative improvement) on internal validation.
Ensembling these refinements with nnU-Net brings every stable configuration to a WL Dice of 0.94, and the SACA variant ensemble achieves the best boundary error (HD95) of 2.92 mm on MU-GLIOMA-POST.

\end{abstract}

\section{Introduction}

Automated volumetric segmentation of multi-modal Magnetic Resonance Imaging (MRI) plays an important role in objectively assessing disease progression in neuro-oncology. After maximal safe surgical resection, tracking changes in the resection cavity and residual tumor over time gives clinicians a reliable way to evaluate treatment response and catch early signs of recurrence~\cite{ellingson2017modified}. 
However, the transition from pre-operative to post-operative complicates the task. Surgery reshapes brain anatomy, triggers non-specific edema, while scans collected across different institutions and hardware introduce strong domain shifts~\cite{holden2024glioblastoma,cepeda2024deep}.

The shift from standard Convolutional Neural Networks 
(CNNs)~\cite{ronneberger2015u} to Vision Transformers (ViTs)~\cite{liu2021swin} 
has improved models' ability to capture long-range spatial dependencies. 
Architectures such as SwinUNETR~\cite{hatamizadeh2021swin,he2023swinunetr} use shifted-window self-attention to model global context,  but their sensitivity to out-of-distribution data remains a bottleneck for clinical translation~\cite{isensee2024nnu}. The majority of current methods rely on region-based  objectives like the Generalized Dice Loss (GDL)~\cite{sudre2017generalised} to handle class imbalance; however, recent works suggest that these formulations exhibit hidden biases and performance collapse when applied to external cohorts with varying intensity distributions~\cite{yeung2022unified}.

In this paper, we run a controlled ablation study along three axes: loss function stability, subspace-aware feature calibration, and intensity normalization, to analyze the collapse seen under multi-site domain shift.
The main contributions of the paper are:
\begin{enumerate}

\item We show that the Generalized Dice Loss (GDL)~\cite{sudre2017generalised} leads to generalization collapse in multi-site post-operative glioma segmentation, and that hybrid Dice-CE objectives~\cite{yeung2022unified} train far more stably across sites.

\item We propose a brain-masked percentile normalization pipeline~\cite{reinhold2019evaluating} combined with voxel-level contrastive learning~\cite{you2022momentum}, which makes features more consistent across clinical sites.

\item We introduce the Subspace-Aware Class Attention (SACA) module, which  re-calibrates feature maps~\cite{hu2018squeeze,oktay2018attention}  by applying shared class-conditioned queries across structured channel subspaces per spatial token, improving ET sensitivity by 
8\% (9.1\% relative improvement) over the \textit{Baseline}.
 
\end{enumerate}

\section{Related Work}
\label{sec:related_work}

\textbf{Glioma segmentation.}
Most existing glioma segmentation methods target pre-operative MRI, where 
models are evaluated under relatively consistent imaging 
conditions~\cite{cepeda2024deep}. These methods don't account for treatment-related  changes such as resection cavities, hemorrhage, and post-radiation inflammation. Holden et al.~\cite{holden2024glioblastoma} highlighted this gap by validating post-operative segmentation on a multicenter cohort of nearly $1000$ patients across $12$ hospitals, where the best model achieved only $61\%$ Dice. The BraTS 
2024 post-treatment challenge~\cite{de20242024} established the first benchmark dedicated to this setting, and introduced the four-class taxonomy that we adopt throughout this work: Enhancing Tumor (ET), Non-enhancing Tumor Core (NETC), Surrounding Non-enhancing FLAIR Hyperintensity (SNFH), and Resection Cavity (RC). Following BraTS reporting conventions, we also report two composite regions—Whole Lesion (WL, the union of all four classes) and Tumor Core (TC, the union of ET and NETC)—which are derived post-hoc from the four-class predictions and are not additional segmentation targets.

\textbf{Segmentation architectures.}
The convolutional encoder-decoder design of U-Net~\cite{ronneberger2015u} is still used in most clinical segmentation systems, but as it relies on local convolution operations, its receptive field is limited, making it hard to capture long-range context. TransUNet~\cite{chen2021transunet} tackled this problem by using a ViT encoder, but its 2D architecture makes it ill-suited for volumetric brain MRI. SwinUNETR~\cite{hatamizadeh2021swin,he2023swinunetr} extended this idea to 3D  data using hierarchical shifted-window attention, and it achieved top performance on BraTS 2021, reaching Dice scores of $0.927$ (WT), $0.876$ (TC), and $0.853$ (ET) on the official test set~\cite{hatamizadeh2021swin}.
 But under domain shift it still struggles with out-of-distribution intensity profiles~\cite{isensee2024nnu}, which we quantify in this study.
We use nnU-Net~\cite{isensee2021nnu} as our ensemble reference, since it auto-configures and stays reliable across diverse datasets.

\textbf{Attention and feature calibration.}
The Squeeze-and-Excitation (SENet) architecture~\cite{hu2018squeeze} recalibrates feature maps through global channel-wise weighting, but applies a single set of weights over the entire unified feature space and does not distinguish between semantically distinct latent groups.
Attention gates~\cite{oktay2018attention} suppress irrelevant skip-connection activations to improve localization, but perform spatial voxel-by-voxel gating rather than disentangling structured channel subspaces. 
The proposed SACA module targets both limitations. It splits the bottleneck into structured subspaces and applies shared class-conditioned queries to each one per spatial token, which lets it recalibrate each tumor phenotype on its own.

\textbf{Loss functions and domain robustness.}
GDL~\cite{sudre2017generalised} handles class imbalance via automatic, volume-based class weighting, which has made it the default objective in most segmentation pipelines. However, its weights are derived from training-set tissue volumes, so they become miscalibrated under domain shift and can trigger generalization collapse. Yeung et al.~\cite{yeung2022unified} showed that hybrid Dice-CE formulations give more stable training across varying class distributions. We build on this by experimenting with both objectives under a controlled multi-site evaluation,  using a dedicated \textit{GDL Variant} to measure how much it degrades under external domain shift relative to our hybrid \textit{DiceWCE} baseline.

\section{Methodology}

Post-operative glioma segmentation is challenging for several reasons. Resection cavities alter brain anatomy, surgical artifacts like blood products create intensity outliers, and acquisition protocols differ across sites which can lead to domain shift across clinical institutions. We build on SwinUNETR~\cite{hatamizadeh2021swin,he2023swinunetr} and study three directions: loss function stability, subspace-aware feature calibration, and intensity normalization.

Instead of the standard GDL, which our \textit{GDL Variant} shows collapses under domain shift, we establish our \textit{Baseline} with a hybrid Dice-CE objective. 

In addition, we introduce the SACA module (\textit{SACA Variant}) to disentangle ambiguous post-surgical signals at the bottleneck, and we pair brain-masked percentile normalization with voxel-level contrastive learning (\textit{Norm Variant}) to make features more consistent across sites. All configurations are then ensembled with an independently trained nnU-Net~\cite{isensee2021nnu}.

\subsection{Datasets, Preprocessing, and Normalization}

In our experiments we use two datasets, so that we can separate in-distribution performance from cross-site generalization: MU-GLIOMA-POST for training and internal validation, and UCSF-ALPTDG, held out from training entirely, to test how the models transfer to a different institution.

\textit{MU-GLIOMA-POST (Internal Cohort):} We chose MU-GLIOMA-POST~\cite{mahmoud2025mu} for its focus on the post-operative four-class taxonomy. After excluding incomplete acquisitions, it comprises $594$ multi-modal scans (T1n, T1c, T2w, FLAIR) from $203$ patients (including longitudinal acquisitions), and we held out $48$ patients for internal validation. The ground truth labels include ET, NETC, SNFH, and RC (Section~\ref{sec:related_work}). On top of these, following the BraTS reporting convention~\cite{de20242024}, we report two composite regions: the Whole Lesion (WL), defined as the union of all four classes, and the Tumor Core (TC), defined as the union of ET and NETC.

\textit{UCSF-ALPTDG (External Validation):} We use the UCSF dataset~\cite{fields2024university} to test external generalization. Evaluating on a $100$-scan subset ($50$ patients) allows us to measure how multi-institutional domain shift and differing acquisition protocols affect performance.

\textit{Preprocessing and Normalization:} Volumes are channel-first aligned, 
RAS-oriented, foreground-cropped ($10$-voxel margin), and padded to 
$160 \times 160 \times 160$. Augmentation includes flipping, rotations, 
intensity shifts, and random patch extraction (2 patches per volume, with an $80/20$ foreground-to-background center-voxel sampling ratio). Following intensity-normalization practice~\cite{reinhold2019evaluating}, we compare two normalization strategies: 
standard channel-wise Z-score on non-zero voxels (\textit{Baseline}, 
\textit{SACA Variant}, \textit{GDL Variant}), and brain-masked percentile clipping 
($1$\textsuperscript{st}--$99$\textsuperscript{th}) followed by Z-score exclusively on valid 
tissues (\textit{Norm Variant}).

\subsection{Baseline Architecture: SwinUNETR}

SwinUNETR~\cite{hatamizadeh2021swin,he2023swinunetr} is a hierarchical 
encoder-decoder that applies shifted-window self-attention over 3D volumetric 
inputs $\mathbf{x} \in \mathbb{R}^{H \times W \times D \times M}$, where $M=4$ 
denotes the number of input modalities, to capture long-range spatial 
dependencies. The encoder progressively merges patch tokens across four stages, 
producing a feature hierarchy with channel dimensions doubling at each stage, 
while skip connections link each encoder stage to its corresponding CNN-based 
decoder block for spatial recovery. The deepest encoder output constitutes the bottleneck representation 
$\mathbf{f} \in \mathbb{R}^{B \times N \times C}$, with $C = 768$ and 
$N = D \times H \times W$ denoting the number of spatial tokens, which 
serves as the primary site of our SACA intervention.

\subsection{Subspace-Aware Class Attention (SACA)}

\begin{figure}[htbp]
\centering
\includegraphics[width=\columnwidth]{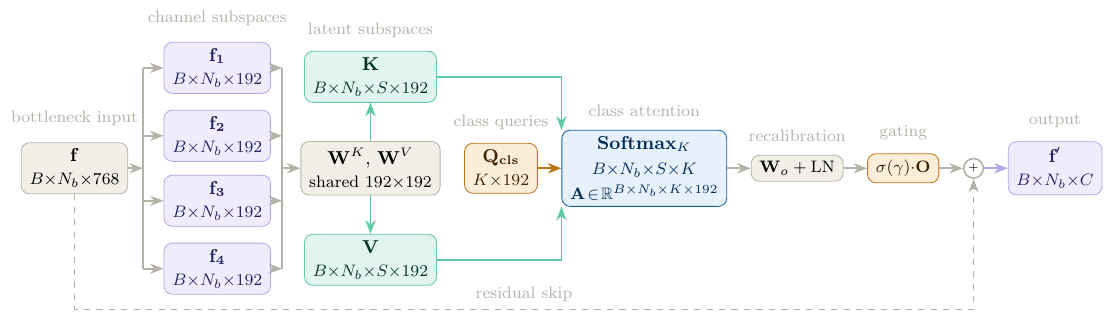}
\caption{Overview of the SACA module. The bottleneck features are partitioned into $S{=}4$ latent subspaces for class-conditioned attention using shared projections and global queries $\mathbf{Q}_{\mathrm{cls}}$. The final output $\mathbf{f}'$ is integrated through a learned gated residual connection: $\mathbf{f}' = \mathbf{f} + \sigma(\gamma) \cdot \mathbf{O}$.}
\label{fig:saca}
\end{figure}

Given the bottleneck representation $\mathbf{f} \in \mathbb{R}^{B \times N \times C}$,
a single recalibration applied to the whole feature space can blur signals that belong to different tumor phenotypes: for instance, the contrast-enhancement of active tumor (ET) against the diffuse hyperintensity of peritumoral edema (SNFH). To keep these signals separate, we split the channel dimension into $S = 4$ non-overlapping subspaces of equal sizes $d_{\mathrm{sub}} = C / S = 192$, setting $S = K = 4$ to match the
number of target segmentation classes. This gives structured feature groups
$\{\mathbf{f}_s\}_{s=1}^{S}$, where $\mathbf{f}_s \in \mathbb{R}^{B \times N
\times d_{\mathrm{sub}}}$. Figure~\ref{fig:saca} gives an overview of the module.

We define $K = 4$ learnable class-specific query vectors $\mathbf{Q}_{\mathrm{cls}} \in \mathbb{R}^{K \times d_{\mathrm{sub}}}$. For each subspace $s$, keys and values come from shared projections $\mathbf{W}^K, \mathbf{W}^V \in
\mathbb{R}^{d_{\mathrm{sub}} \times d_{\mathrm{sub}}}$, while $\mathbf{Q}_{\mathrm{cls}}$
is learned directly without projection, giving intermediate attention weights of
shape $B \times N \times S \times K$.
The module then \textit{contracts} subspaces onto classes: for each spatial token $n$, we compute scaled dot-product attention between the class queries and the per-subspace keys, with softmax over the $K$ class dimension, and sum the weighted values across all $S$ subspaces. This collapses into class-calibrated features $\mathbf{A}^{(n)} \in \mathbb{R}^{K \times d_{\mathrm{sub}}}$:

\[
\mathbf{A}^{(n)} = \sum_{s=1}^S \mathrm{Softmax}_K \!\left(
    \frac{\mathbf{Q}_{\mathrm{cls}}\,(\mathbf{K}_s^{(n)})^\top}{\sqrt{d_{\mathrm{sub}}}}
\right) \mathbf{V}_s^{(n)}
\]

We then concatenate the $K$ class-specific representations $\mathbf{A}^{(n)}$ 
along the class axis across all spatial tokens, which restores the original channel depth ($K \times d_{\mathrm{sub}} = C = 768$), and project them with
$\mathbf{W}_o \in \mathbb{R}^{C \times C}$. The output  enters through a gated residual, where $\gamma \in \mathbb{R}$, initialized to $0$, is passed through a sigmoid $\sigma(\cdot)$, so the module starts as an identity map and gains influence only as training proceeds:

\[
\mathbf{f}' = \mathbf{f} + \sigma(\gamma) \cdot
    \mathrm{LN}\!\left(
        \bigl[\mathbf{A}_{:,:,1,:} \,\|\, \cdots \,\|\,
        \mathbf{A}_{:,:,K,:}\bigr]\,\mathbf{W}_o
    \right)
\]

\subsection{Loss Function Ablation: GDL vs.\ DiceWCE}

A main hypothesis of this study is that the standard generalized dice loss 
(GDL)~\cite{sudre2017generalised} becomes unstable under external domain shift. The \textit{GDL Variant} tests this directly: it trains the model using standard GDL. The other variants (\textit{Baseline}, \textit{SACA Variant},  \textit{Norm Variant}) instead use a manually weighted hybrid loss (\textit{DiceWCE}) for more stable convergence.

DiceWCE combines Dice Loss (squared predictions, excluding background) with a 
class-weighted Cross-Entropy (CE) objective~\cite{yeung2022unified}, where the active-tumor subclasses get higher weights so the model prioritizes residual tissue.

The segmentation loss is defined as:
$\mathcal{L}_{\text{seg}} = \mathcal{L}_{\text{Dice}} + 0.6 \cdot 
\mathcal{L}_{\text{CE}}$.

\subsection{Voxel-Level Contrastive Learning (Norm Variant)}

In the \textit{Norm Variant}, we couple the percentile normalization with a voxel-level 
contrastive loss ($\mathcal{L}_{\text{contra}}$)~\cite{you2022momentum} 
applied at the bottleneck to push the features toward being hardware-invariant. Features are projected through a two-layer MLP 
$\phi: \mathbb{R}^{768} \rightarrow \mathbb{R}^{128}$ and $\ell_2$-normalized.

We build the contrastive pairs around tissues that are easy to confuse:
for example, active tumor voxels (ET) serve as positive anchors 
and are contrasted against background and edema (SNFH). We sample up 
to $64$ positive and $128$ negative voxels per pair, using a temperature scaling 
factor of $\tau=0.5$.

The contrastive weight $\alpha_{\text{contra}}$ is defined as a function of 
the training epoch $e$:
\[
\alpha_{\text{contra}}(e) = 
\begin{cases} 
0 & \text{if } e \leq e_{\text{warm}} \\ 
0.15 \cdot \dfrac{e - e_{\text{warm}}}{e_{\text{ramp}}} & \text{if } e > e_{\text{warm}}
\end{cases}
\]
where $e_{\text{warm}}$ and $e_{\text{ramp}}$ denote the warmup and 
ramp-up durations, respectively. The total training objective is then: $
\mathcal{L}_{\text{total}} = \mathcal{L}_{\text{seg}} + 
\alpha_{\text{contra}}(e) \cdot \mathcal{L}_{\text{contra}}$.

\subsection{Optimization and Ensemble Strategy}

All models were optimized using AdamW ($\text{lr} = 2 \times 10^{-4}$, $\beta_1 = 0.9$, $\beta_2 = 0.999$, weight decay $= 3 \times 10^{-5}$) for $500$ epochs. The learning rate schedule consisted of a $10$-epoch linear warmup from $2 \times 10^{-6}$ to $2 \times 10^{-4}$, followed by cosine annealing over the remaining $490$ epochs to a minimum of $1 \times 10^{-6}$.

For the final segmentations, we ensemble each SwinUNETR model with an independently trained 3D nnU-Net~\cite{isensee2021nnu} by averaging their voxel-wise softmax probabilities~\cite{lakshminarayanan2017simple}.

\section{Results}

We organize the evaluation as follows: (i) standalone performance on the internal MU-GLIOMA-POST dataset, (ii) robustness under domain shift using the external UCSF-ALPTDG validation set, (iii) the effect of the SACA Variant on detection sensitivity, and (iv) the gains from ensembling with nnU-Net.

\subsection{Standalone Performance and Internal Validation}

Table~\ref{tab:mu_glioma_results} reports standalone performance on the internal validation set ($n = 48$), with an independently trained 
nnU-Net~\cite{isensee2021nnu} as the reference 
(WL Dice $0.94 \pm 0.15$). Among SwinUNETR configurations, the \textit{Norm Variant} achieved the highest WL mean Dice ($0.92 \pm 0.15$), while the \textit{SACA Variant} obtained comparable overlap ($0.92 \pm 0.18$) and the best boundary fidelity (median WL HD95: $2.12$ mm). The \textit{GDL Variant} was the least stable: its WL Dice dropped to $0.88 \pm 0.16$ and HD95 deteriorated to $11.48 \pm 17.29$ mm, a $2.7$-fold error increase over the \textit{Baseline}.

\begin{table}[!t]
\centering
\caption{Internal validation results on MU-GLIOMA-POST ($n = 48$). Performance metrics include Dice Similarity Coefficient (DSC) and 95\textsuperscript{th} percentile Hausdorff Distance (HD95).
}
\label{tab:mu_glioma_results}
\renewcommand{\arraystretch}{1.2}
\adjustbox{max width=\textwidth}{%
\begin{tabular}{ll cc cc}
\toprule
\multirow{2}{*}{\textbf{Region}}
& \multirow{2}{*}{\textbf{Model}}
& \multicolumn{2}{c}{\textbf{Dice Score $\uparrow$}}
& \multicolumn{2}{c}{\textbf{HD95 (mm) $\downarrow$}} \\
\cmidrule(lr){3-4}\cmidrule(lr){5-6}
& & \textbf{Mean $\pm$ SD} & \textbf{Median (IQR)}
& \textbf{Mean $\pm$ SD} & \textbf{Median (IQR)} \\
\midrule
\multirow{5}{*}{WL}
& nnU-Net
& $0.94 \pm 0.15$ & $0.98$ ($0.97$--$0.99$)
& $3.03 \pm 5.49$ & $1.00$ ($1.00$--$2.24$) \\
& Baseline
& $0.91 \pm 0.19$ & $\mathbf{0.96}$ (\textbf{0.94}--\textbf{0.97})
& $\mathbf{4.32 \pm 6.19}$ & $2.24$ ($1.41$--$4.12$) \\
& GDL Variant
& $0.88 \pm 0.16$ & $0.92$ ($0.88$--$0.94$)
& $11.48 \pm 17.29$ & $6.12$ ($3.32$--$10.14$) \\
& Norm Variant
& $\mathbf{0.92 \pm 0.15}$ & $\mathbf{0.96}$ (\textbf{0.93}--\textbf{0.97})
& $5.11 \pm 10.16$ & $2.24$ ($1.41$--$4.03$) \\
& SACA Variant
& $\mathbf{0.92 \pm 0.18}$ & $\mathbf{0.96}$ (\textbf{0.94}--\textbf{0.97})
& $5.40 \pm 11.93$ & $\mathbf{2.12}$ (\textbf{1.41}--\textbf{3.70}) \\
\midrule
\multirow{5}{*}{TC}
& nnU-Net
& $0.87 \pm 0.17$ & $0.92$ ($0.84$--$0.96$)
& $3.20 \pm 4.19$ & $1.41$ ($1.00$--$3.90$) \\
& Baseline
& $\mathbf{0.83 \pm 0.17}$ & $\mathbf{0.90}$ (\textbf{0.78}--\textbf{0.94})
& $\mathbf{4.78 \pm 5.17}$ & $\mathbf{3.00}$ (\textbf{1.41}--\textbf{5.60}) \\
& GDL Variant
& $0.82 \pm 0.18$ & $0.87$ ($0.80$--$0.92$)
& $5.09 \pm 4.84$ & $3.46$ ($2.24$--$5.87$) \\
& Norm Variant
& $0.78 \pm 0.19$ & $0.81$ ($0.72$--$0.91$)
& $6.15 \pm 6.10$ & $4.18$ ($2.24$--$7.08$) \\
& SACA Variant
& $0.82 \pm 0.19$ & $0.89$ ($0.79$--$0.94$)
& $5.19 \pm 6.96$ & $3.17$ ($1.41$--$5.10$) \\
\midrule
\multirow{5}{*}{ET}
& nnU-Net
& $0.89 \pm 0.20$ & $0.94$ ($0.90$--$0.97$)
& $2.74 \pm 4.35$ & $1.00$ ($1.00$--$1.73$) \\
& Baseline
& $\mathbf{0.84 \pm 0.21}$ & $\mathbf{0.91}$ (\textbf{0.83}--\textbf{0.95})
& $4.27 \pm 5.50$ & $\mathbf{1.41}$ (\textbf{1.00}--\textbf{4.81}) \\
& GDL Variant
& $0.83 \pm 0.20$ & $0.90$ ($0.82$--$0.93$)
& $\mathbf{4.25 \pm 5.10}$ & $2.00$ ($1.41$--$4.68$) \\
& Norm Variant
& $0.80 \pm 0.20$ & $0.87$ ($0.76$--$0.92$)
& $5.56 \pm 6.33$ & $2.45$ ($1.41$--$6.08$) \\
& SACA Variant
& $0.83 \pm 0.21$ & $\mathbf{0.91}$ (\textbf{0.81}--\textbf{0.95})
& $4.61 \pm 7.20$ & $1.73$ ($1.00$--$4.26$) \\
\bottomrule
\end{tabular}}
\end{table}

\subsection{Generalization Gap and Domain Shift (UCSF-ALPTDG)}

To test translational robustness, we evaluated all models on the external 
UCSF-ALPTDG test set ($n = 100$), using the official nnU-Net~\cite{isensee2021nnu} 
results from~\cite{fields2024university} as an in-distribution reference. 
As shown in Table~\ref{tab:ucsf_results}, every architecture lost some accuracy on the external set: the \textit{Baseline} WL Dice fell by about 
$7$ percentage points relative to the internal results.  This gap reflects the domain shift caused by different scanners and acquisition protocols across institutions.

\textbf{GDL collapse.} The \textit{GDL Variant} was by far the least stable under this shift. Its WL Dice dropped from 
$0.88$ to $0.73$, a decrease of $14.7$~pp, and its WL HD95 increased from $11.48$ mm internally to  $36.31$ mm externally (a $3.2\times$ increase). The peritumoral edema region (SNFH) showed the same pattern. This is consistent with GDL's volume-based class weighting overfitting to the training site's tissue proportions and failing to transfer to different hardware.

\begin{table}[htbp]
\centering
\caption{External validation results on the UCSF-ALPTDG test set ($n = 100$). Performance metrics include Dice Similarity Coefficient (DSC) and Whole Lesion 95\textsuperscript{th} percentile Hausdorff Distance (WL HD95). nnU-Net~\cite{isensee2021nnu} serves as the in-distribution reference benchmark, with baseline scores adopted from~\cite{fields2024university}.}
\label{tab:ucsf_results}
\renewcommand{\arraystretch}{1.2}
\adjustbox{max width=\textwidth}{%
\begin{tabular}{l cccccc c}
\toprule
\multirow{2}{*}{\textbf{Model}} & \multicolumn{6}{c}{\textbf{Dice Score $\uparrow$}} & \textbf{HD95 (mm) $\downarrow$} \\
\cmidrule(lr){2-7} \cmidrule(lr){8-8}
& \textbf{WL} & \textbf{TC} & \textbf{ET} & \textbf{NETC} & \textbf{RC} & \textbf{SNFH} & \textbf{WL} \\
\midrule
nnU-Net & $0.87 \pm 0.17$ & $0.75 \pm 0.25$ & $0.76 \pm 0.25$ & $0.53 \pm 0.30$ & $0.76 \pm 0.28$ & $0.83 \pm 0.18$ & $8.93 \pm 18.84$ \\
\midrule
Baseline & $\mathbf{0.84 \pm 0.17}$ & $0.71 \pm 0.24$ & $0.73 \pm 0.25$ & $0.46 \pm 0.28$ & $\mathbf{0.59 \pm 0.36}$ & $\mathbf{0.82 \pm 0.17}$ & $\mathbf{10.04 \pm 16.48}$ \\
GDL Variant & $0.73 \pm 0.21$ & $0.69 \pm 0.24$ & $0.71 \pm 0.24$ & $0.38 \pm 0.30$ & $0.45 \pm 0.36$ & $0.72 \pm 0.20$ & $36.31 \pm 26.04$ \\
Norm Variant & $0.83 \pm 0.18$ & $\mathbf{0.73 \pm 0.23}$ & $0.73 \pm 0.23$ & $\mathbf{0.54 \pm 0.26}$ & $\mathbf{0.59 \pm 0.35}$ & $0.81 \pm 0.17$ & $13.61 \pm 17.24$ \\
SACA Variant & $0.83 \pm 0.18$ & $\mathbf{0.73 \pm 0.24}$ & $\mathbf{0.74 \pm 0.24}$ & $0.45 \pm 0.32$ & $\mathbf{0.59 \pm 0.37}$ & $0.80 \pm 0.17$ & $15.05 \pm 17.73$ \\
\bottomrule
\end{tabular}}
\end{table}

\subsection{Improving Active Tumor Recall with SACA}

\begin{figure}[htbp]
\centering
\includegraphics[width=0.85\textwidth]{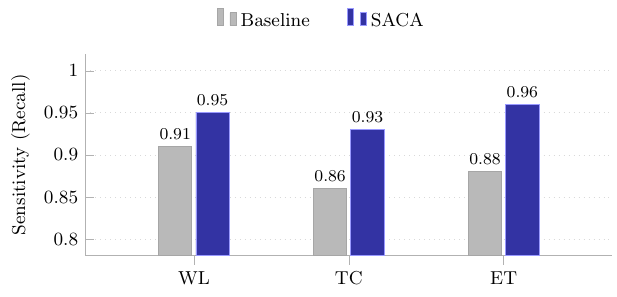}
\caption{Sensitivity (recall) analysis across tumor-active regions. The bar chart illustrates the systematic improvement in detection performance provided by the \textit{SACA Variant} compared to the \textit{Baseline} SwinUNETR.}
\label{fig:sensitivity}

\end{figure}

In post-operative monitoring, sensitivity matters more: a missed enhancing tumor voxel can mean undetected recurrence and delay treatment, while a false positive usually leads to additional follow-up imaging. 
We report sensitivity separately from the aggregate Dice score. Figure~\ref{fig:sensitivity} shows that the \textit{SACA Variant} 
improves recall across tumor-active regions on the internal MU-GLIOMA-POST dataset, increasing ET and TC sensitivity by $8\%$ (9.1\% relative improvement) and $7\%$ ($8.1\%$ relative improvement), respectively. This reflects the module's ability to recalibrate each subspace towards its target class, which helps spot residual tissue.

\subsection{Ensemble Synergy}

We ensembled each variant with nnU-Net~\cite{isensee2021nnu}. As shown in Table~\ref{tab:ensemble_results},
this yielded consistent gains across all stable variants. The \textit{Baseline},
\textit{Norm Variant}, and \textit{SACA Variant} ensembles all converged to an
optimal internal WL Dice of $0.94$, with the \textit{Norm Variant} being the most consistent ($\pm 0.14$). The \textit{SACA Variant} ensemble gave the best boundary fidelity, with a WL HD95 of $2.92$ mm, slightly surpassing
standalone nnU-Net ($3.03$ mm). The \textit{GDL Variant} ensemble showed a  negative result: it pulled nnU-Net down from
$0.94$ to $0.90$ WL Dice and from $3.03$ mm to $7.70$ mm HD95. This could be explained by the systematic prediction bias induced by GDL's volume-based
weighting, which corrupts the stronger model's predictions.

\begin{table}[htbp]
\centering
\caption{Ensemble results on the MU-GLIOMA-POST dataset ($n = 48$). Performance
metrics include Dice Similarity Coefficient (DSC) and Whole Lesion
95\textsuperscript{th} percentile Hausdorff Distance (WL HD95). 
}
\label{tab:ensemble_results}
\renewcommand{\arraystretch}{1.2}
\adjustbox{max width=\textwidth}{%
\begin{tabular}{l ccc c}
\toprule
\multirow{2}{*}{\textbf{Model Configuration}}
& \multicolumn{3}{c}{\textbf{Dice Score $\uparrow$}} & \textbf{HD95 (mm) $\downarrow$} \\
\cmidrule(lr){2-4} \cmidrule(lr){5-5}
& \textbf{WL} & \textbf{TC} & \textbf{ET} & \textbf{WL} \\
\midrule
nnU-Net
& $0.94 \pm 0.15$ & $0.87 \pm 0.17$ & $0.89 \pm 0.20$ & $3.03 \pm 5.49$ \\
\midrule
nnU-Net + Baseline
& $\mathbf{0.94 \pm 0.16}$ & $\mathbf{0.87 \pm 0.17}$ & $0.88 \pm 0.20$
& $3.15 \pm 5.52$ \\
nnU-Net + GDL Variant
& $0.90 \pm 0.16$ & $0.85 \pm 0.18$ & $0.86 \pm 0.20$ & $7.70 \pm 15.68$ \\
nnU-Net + Norm Variant
& $\mathbf{0.94 \pm 0.14}$ & $\mathbf{0.87 \pm 0.17}$ & $0.88 \pm 0.20$
& $3.08 \pm 5.41$ \\
nnU-Net + SACA Variant
& $\mathbf{0.94 \pm 0.15}$ & $\mathbf{0.87 \pm 0.17}$ & $\mathbf{0.89 \pm 0.20}$
& $\mathbf{2.92 \pm 5.17}$ \\
\bottomrule
\end{tabular}}
\vspace{-5pt}
\end{table}

\begin{figure}[!ht]
\centering
\setlength{\tabcolsep}{2pt}
\begin{tabular}{cccccc}
& \small T1c & \small GT & \small Baseline
& \small Norm Variant & \small SACA Variant \\
\rotatebox{90}{\small \hspace{8pt} Case 1} &
\includegraphics[width=0.18\textwidth]{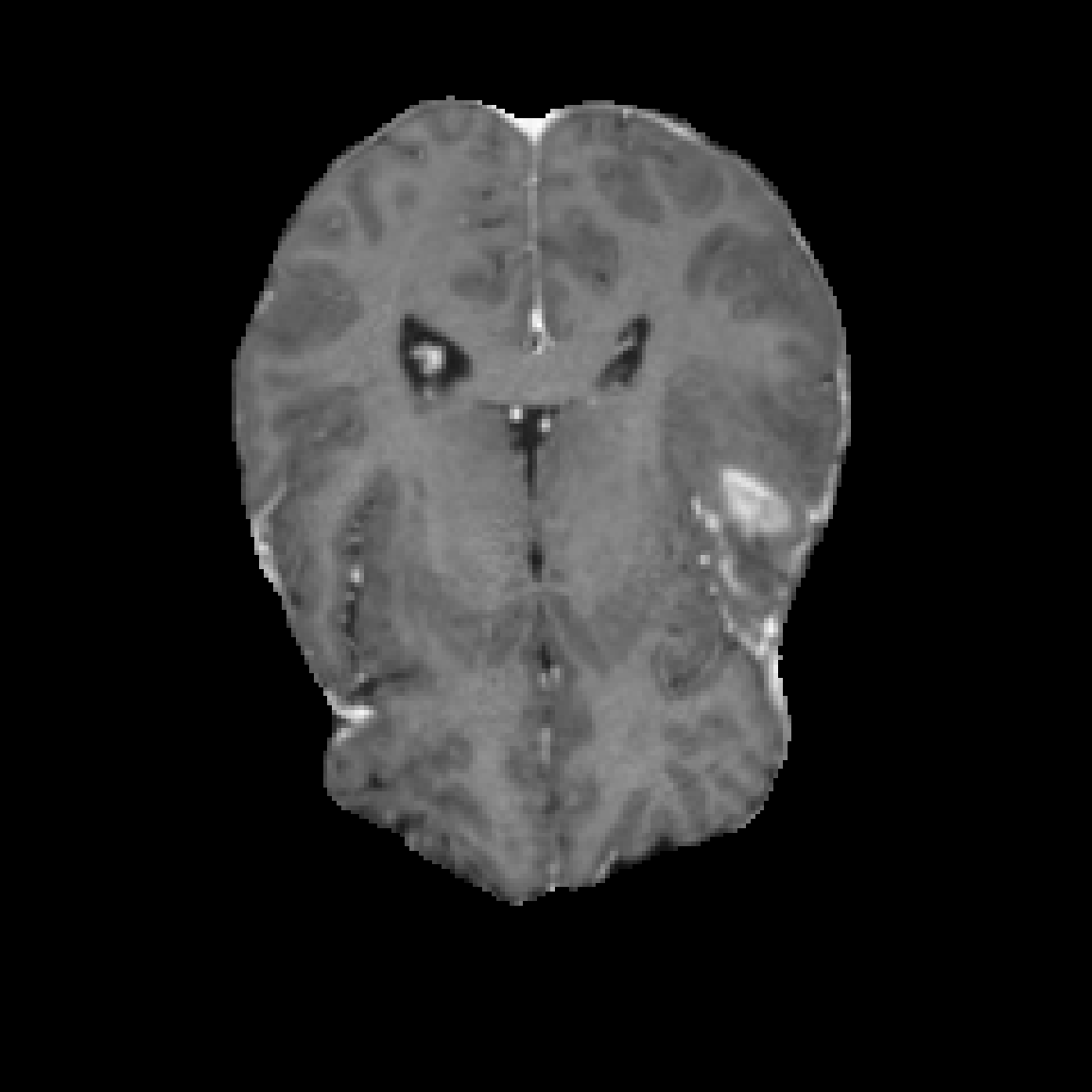} &
\includegraphics[width=0.18\textwidth]{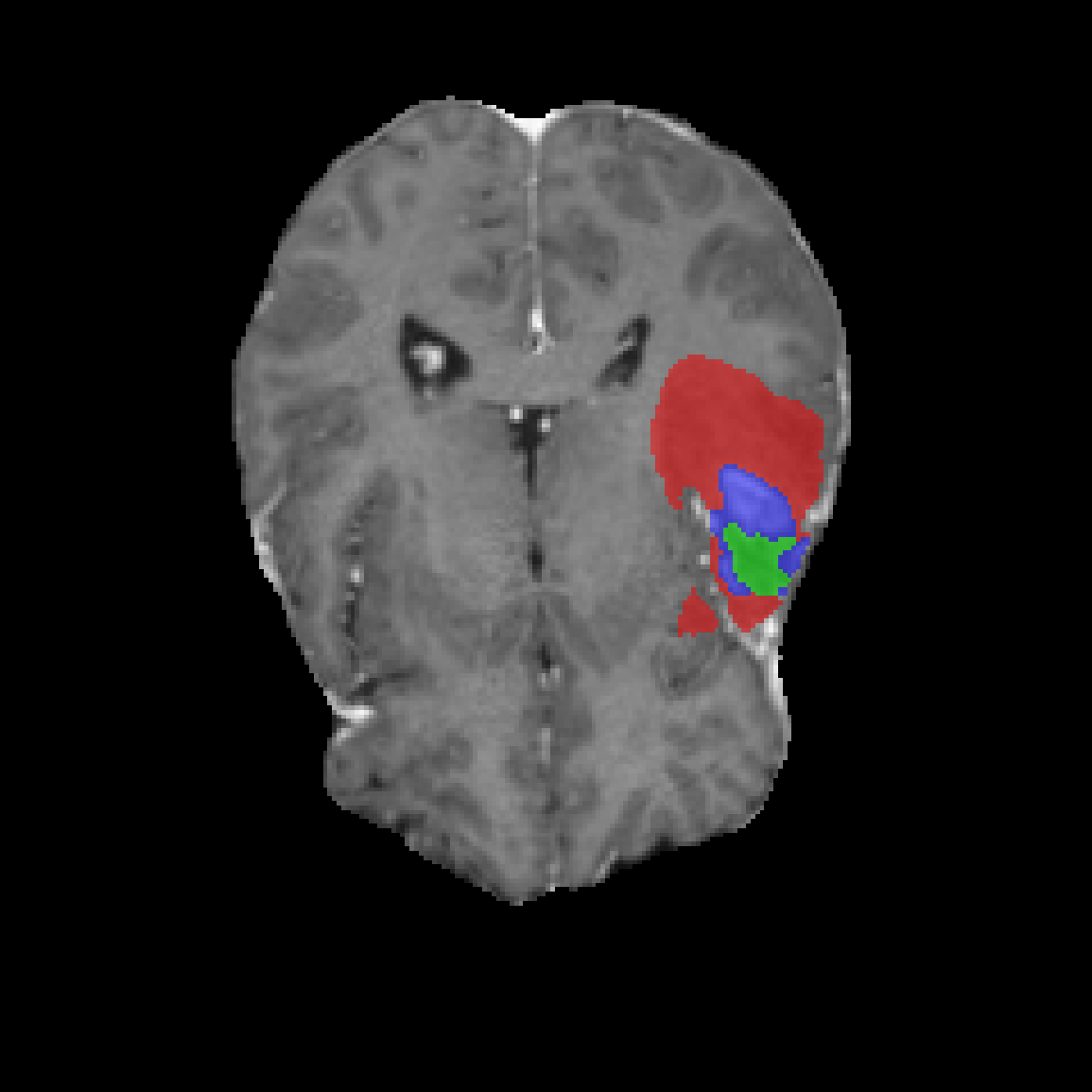} &
\includegraphics[width=0.18\textwidth]{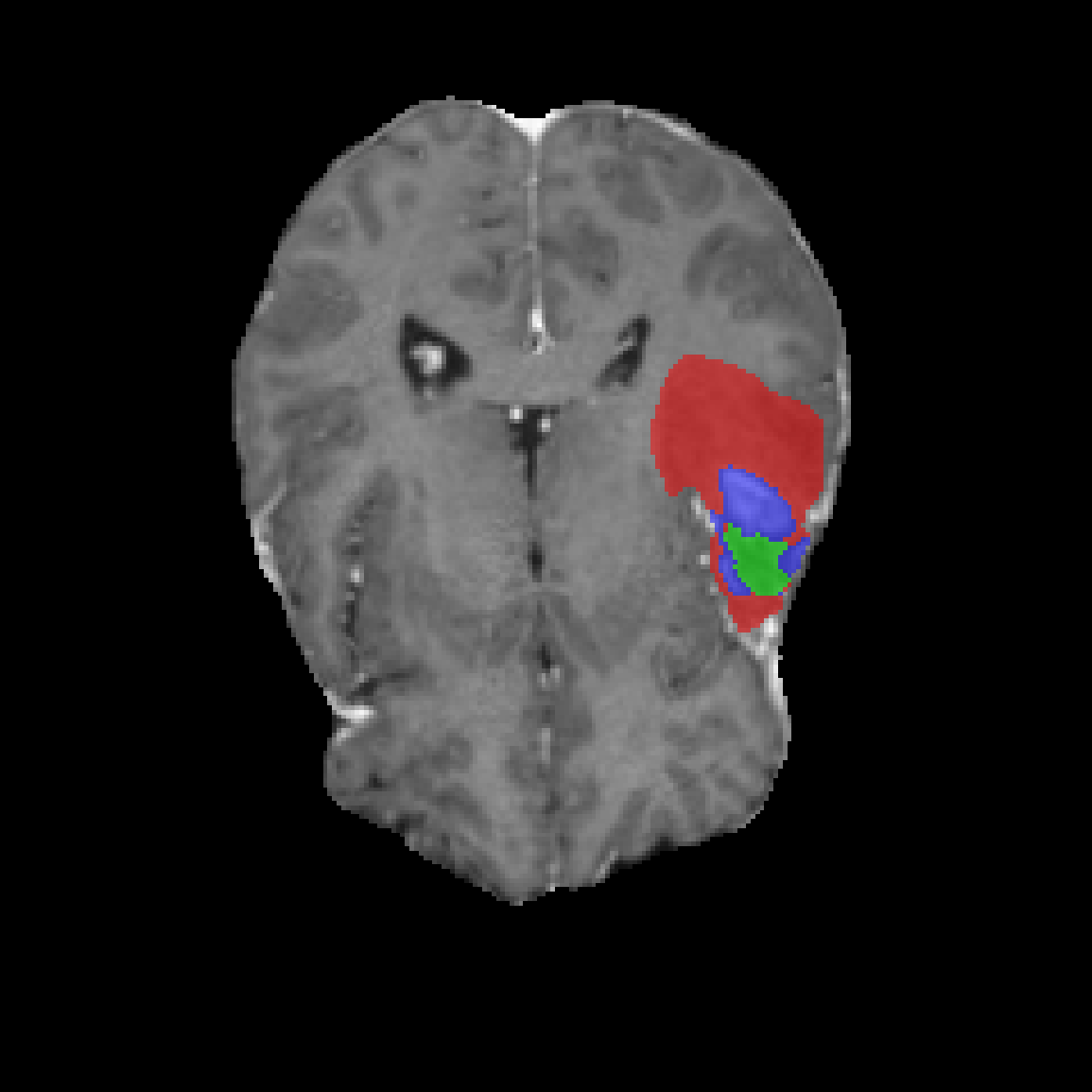} &
\includegraphics[width=0.18\textwidth]{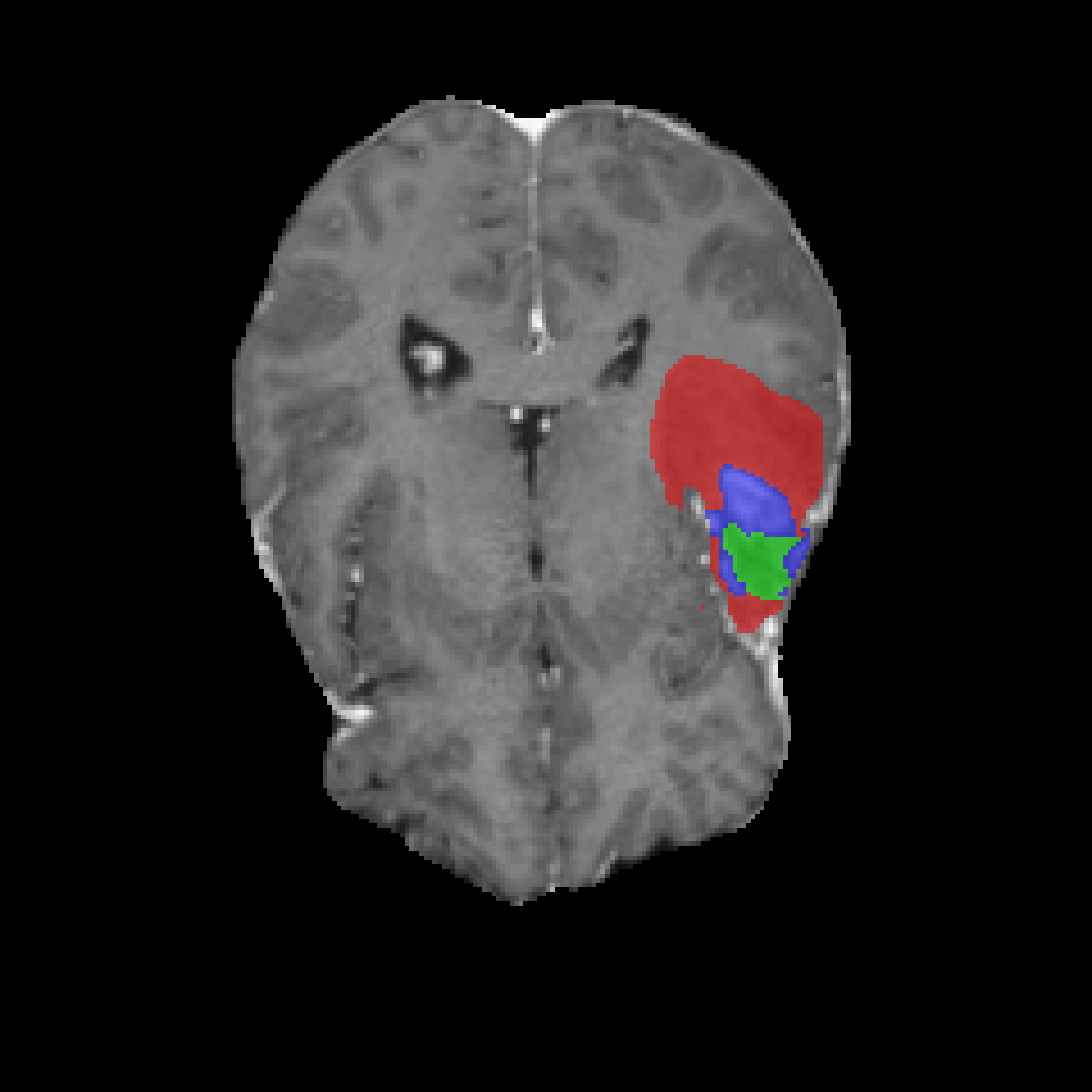} &
\includegraphics[width=0.18\textwidth]{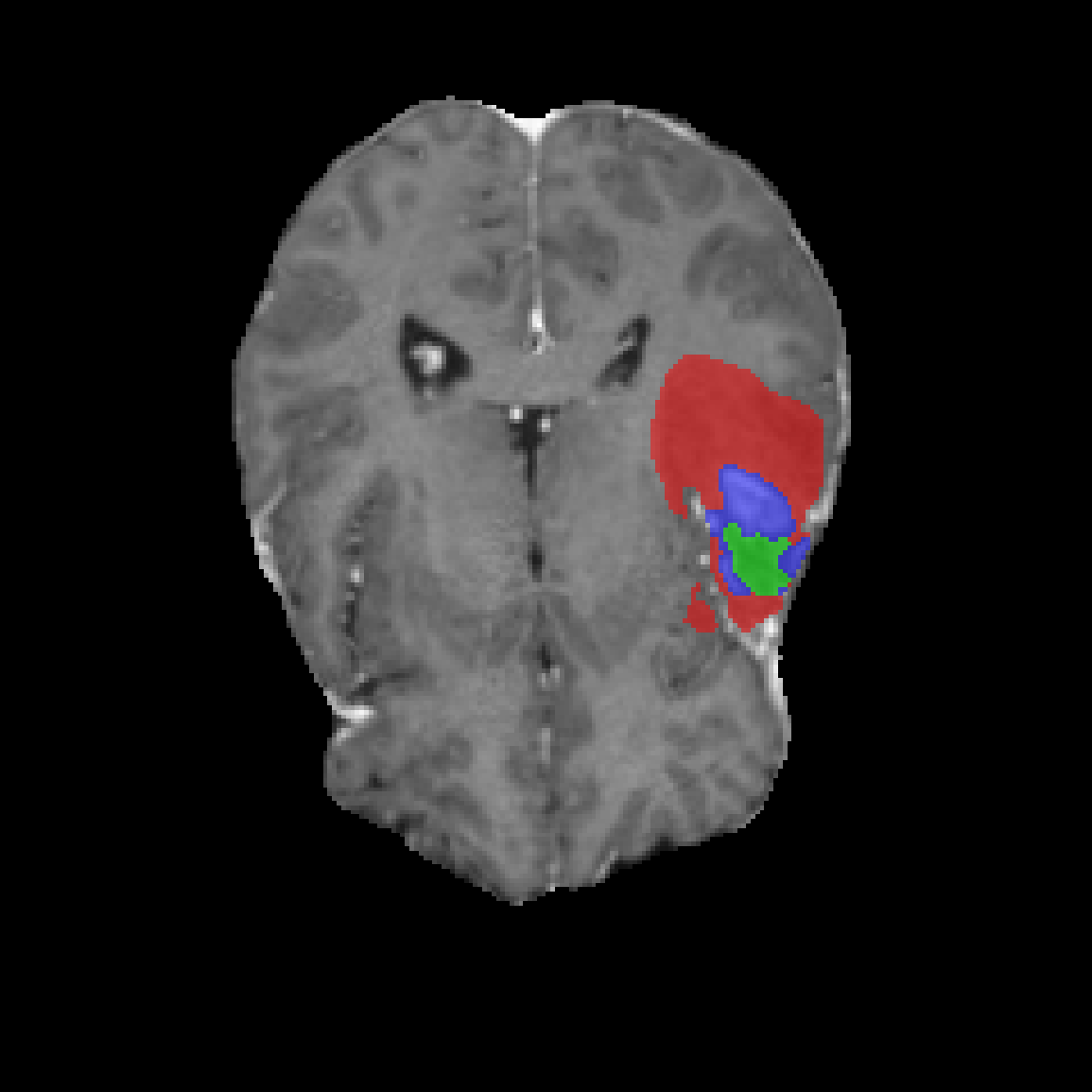} \\[2pt]
\rotatebox{90}{\small \hspace{8pt} Case 2} &
\includegraphics[width=0.18\textwidth]{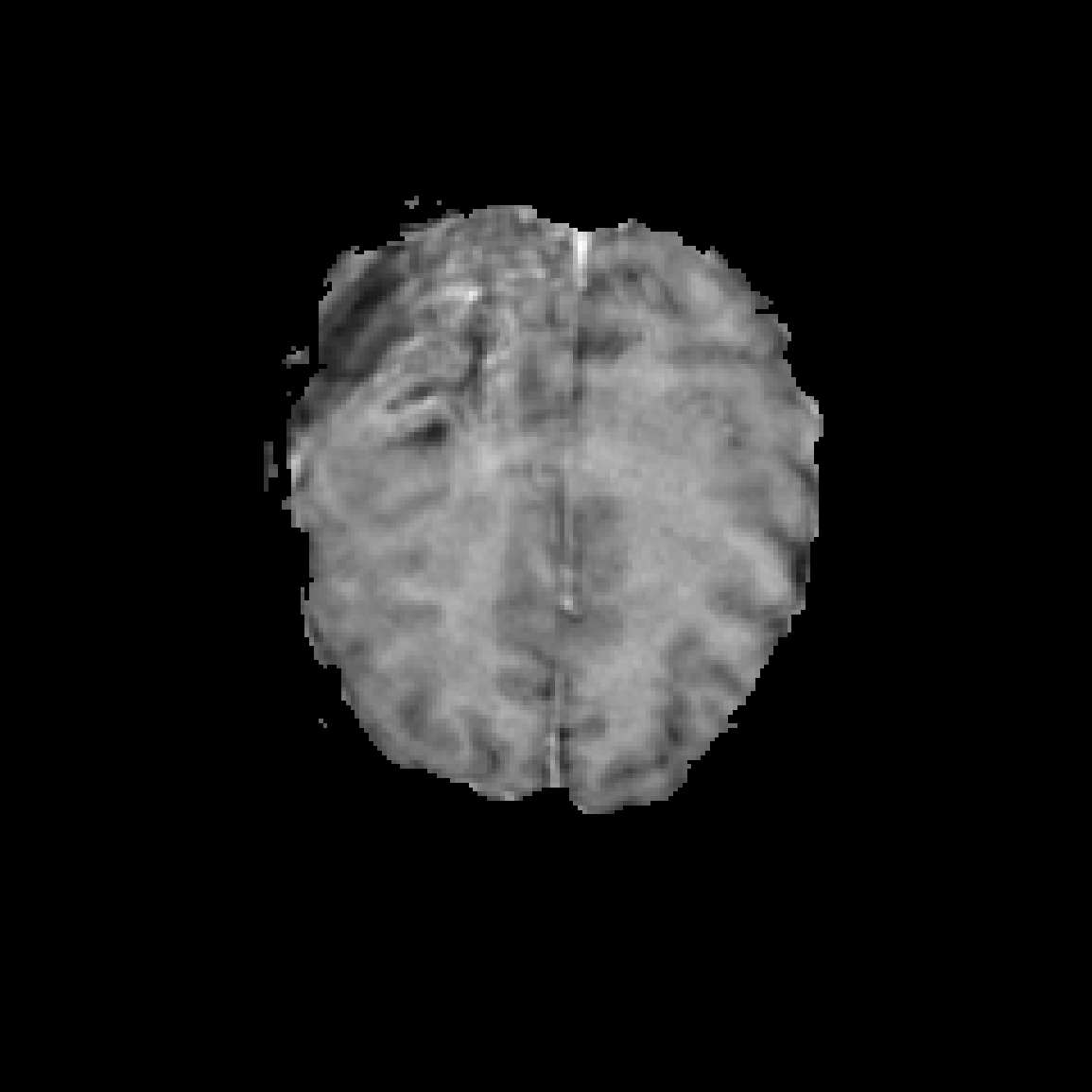} &
\includegraphics[width=0.18\textwidth]{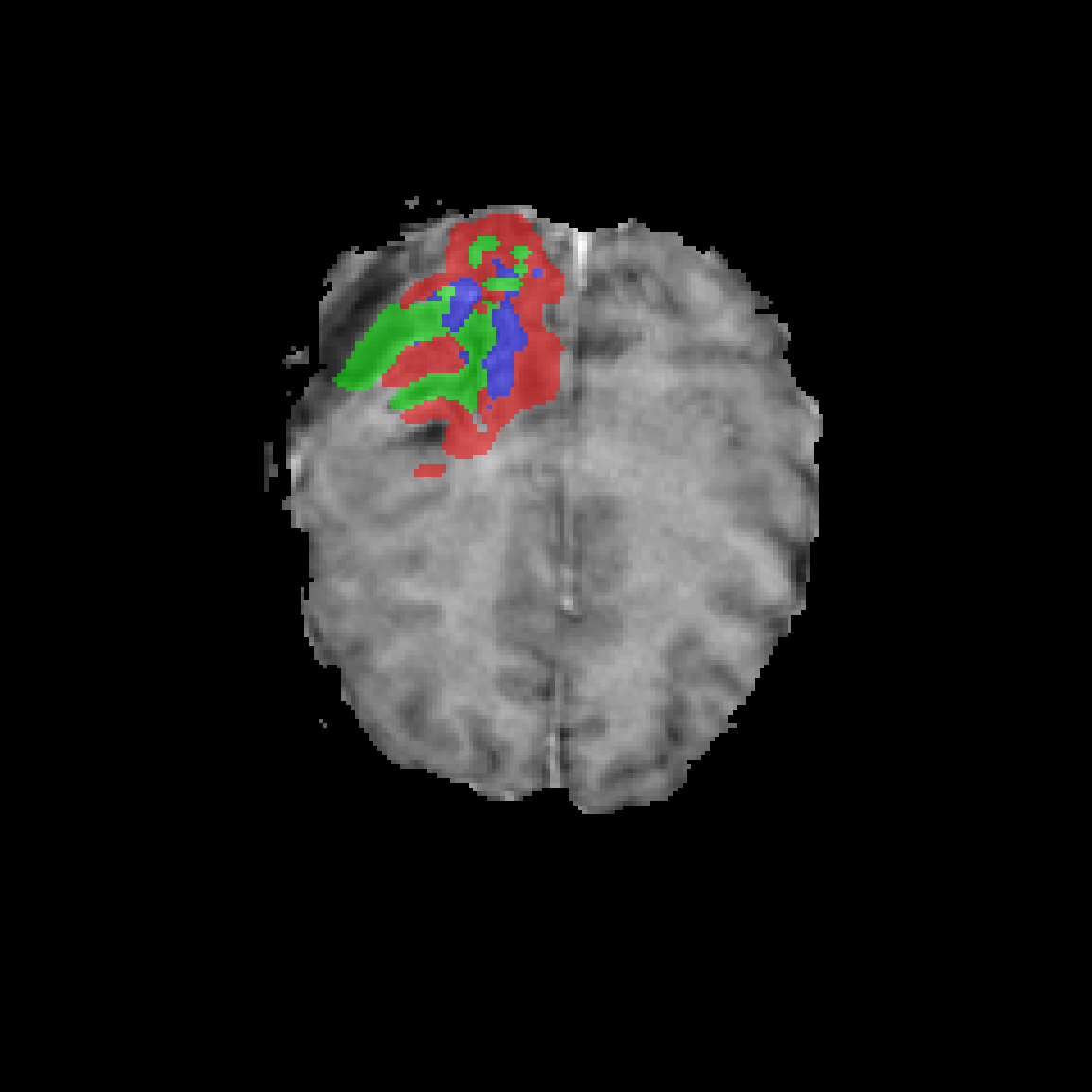} &
\includegraphics[width=0.18\textwidth]{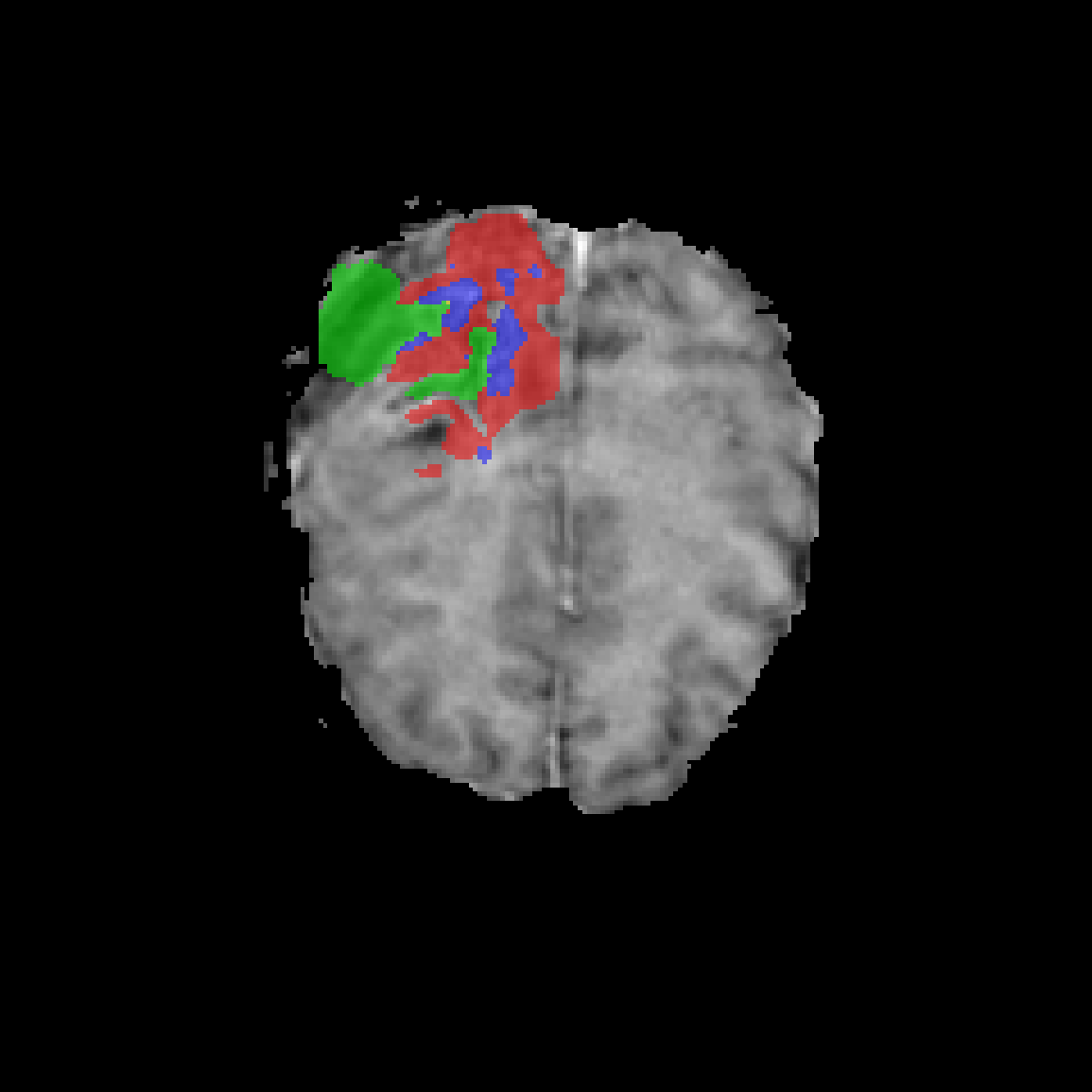} &
\includegraphics[width=0.18\textwidth]{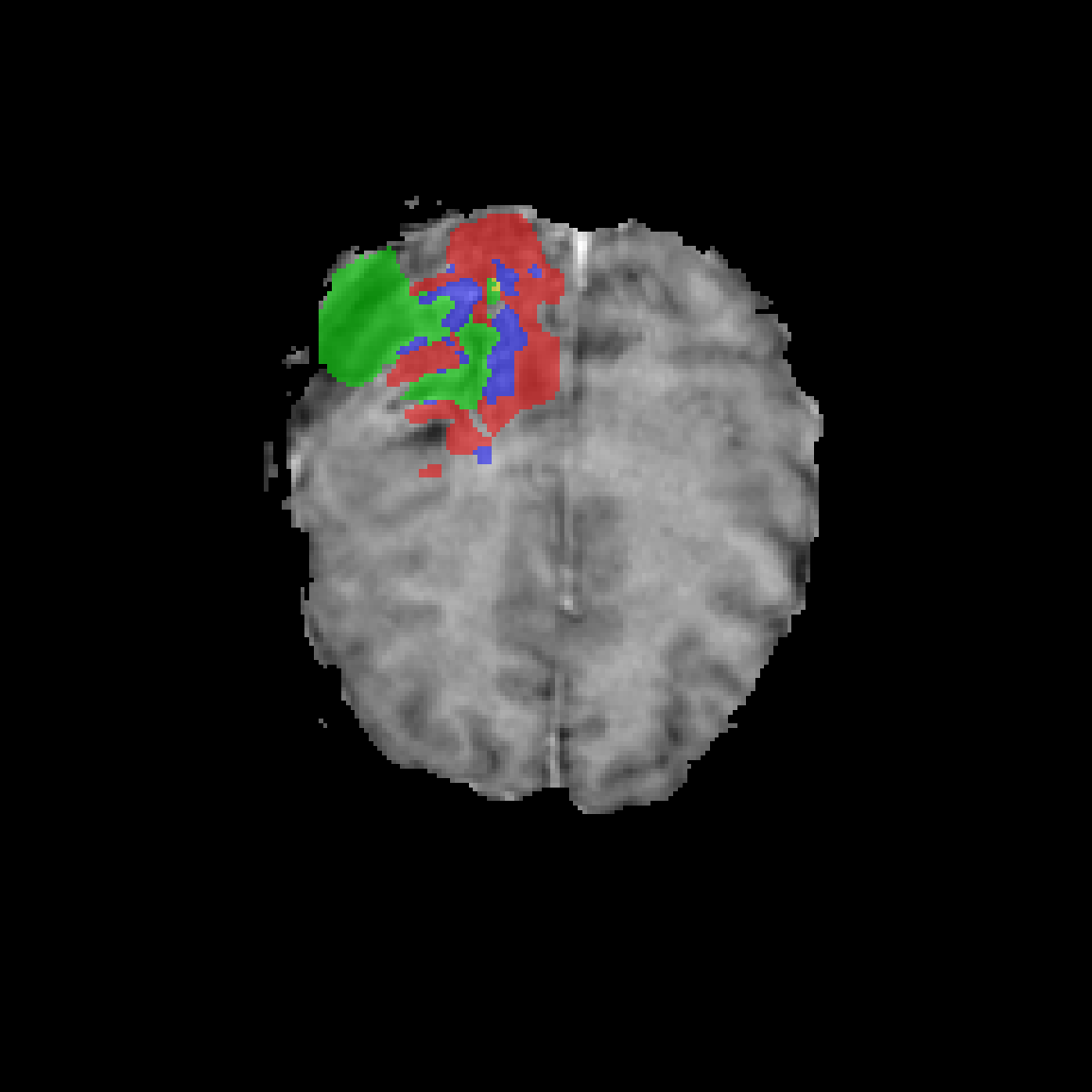} &
\includegraphics[width=0.18\textwidth]{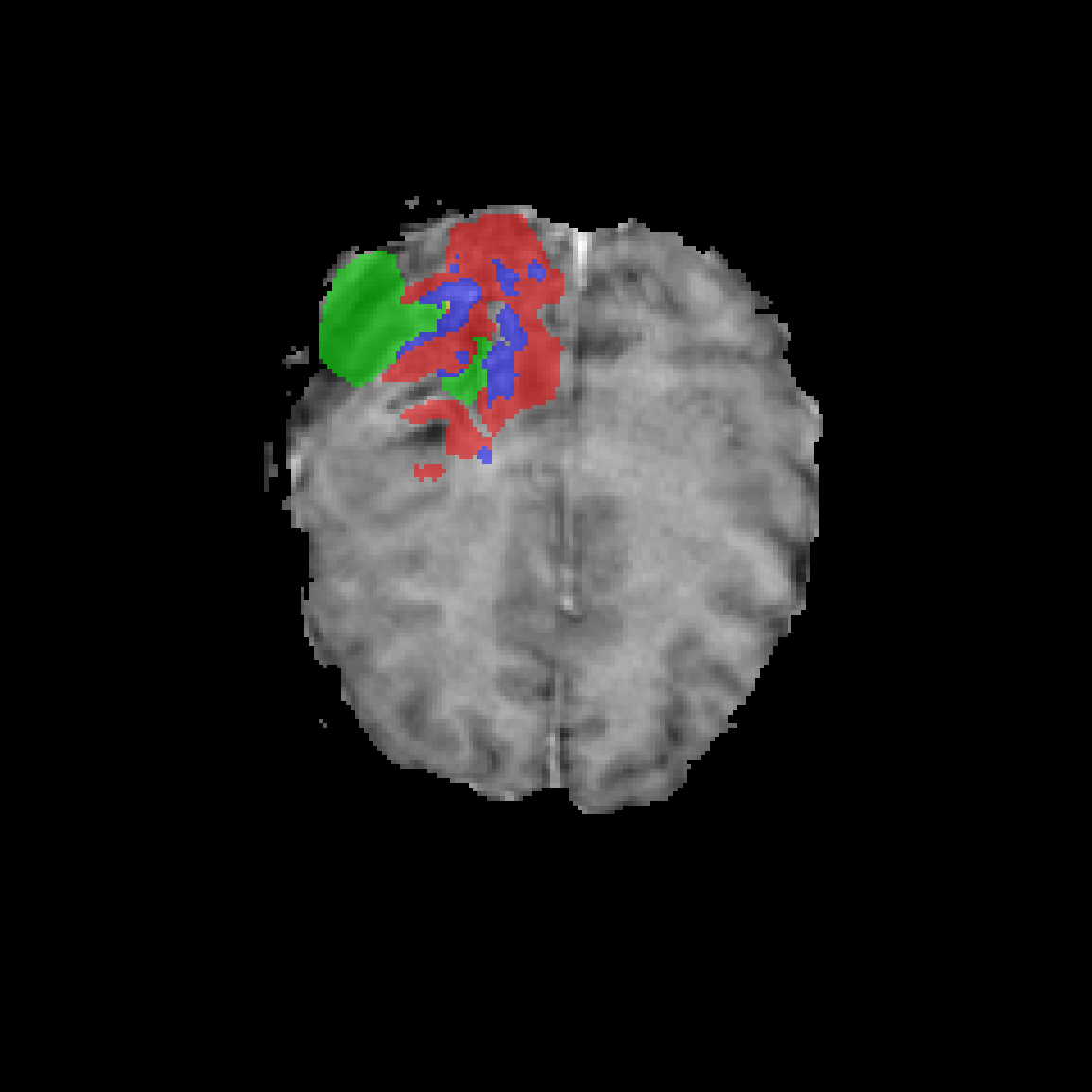} \\[2pt]
\rotatebox{90}{\small \hspace{8pt} Case 3} &
\includegraphics[width=0.18\textwidth]{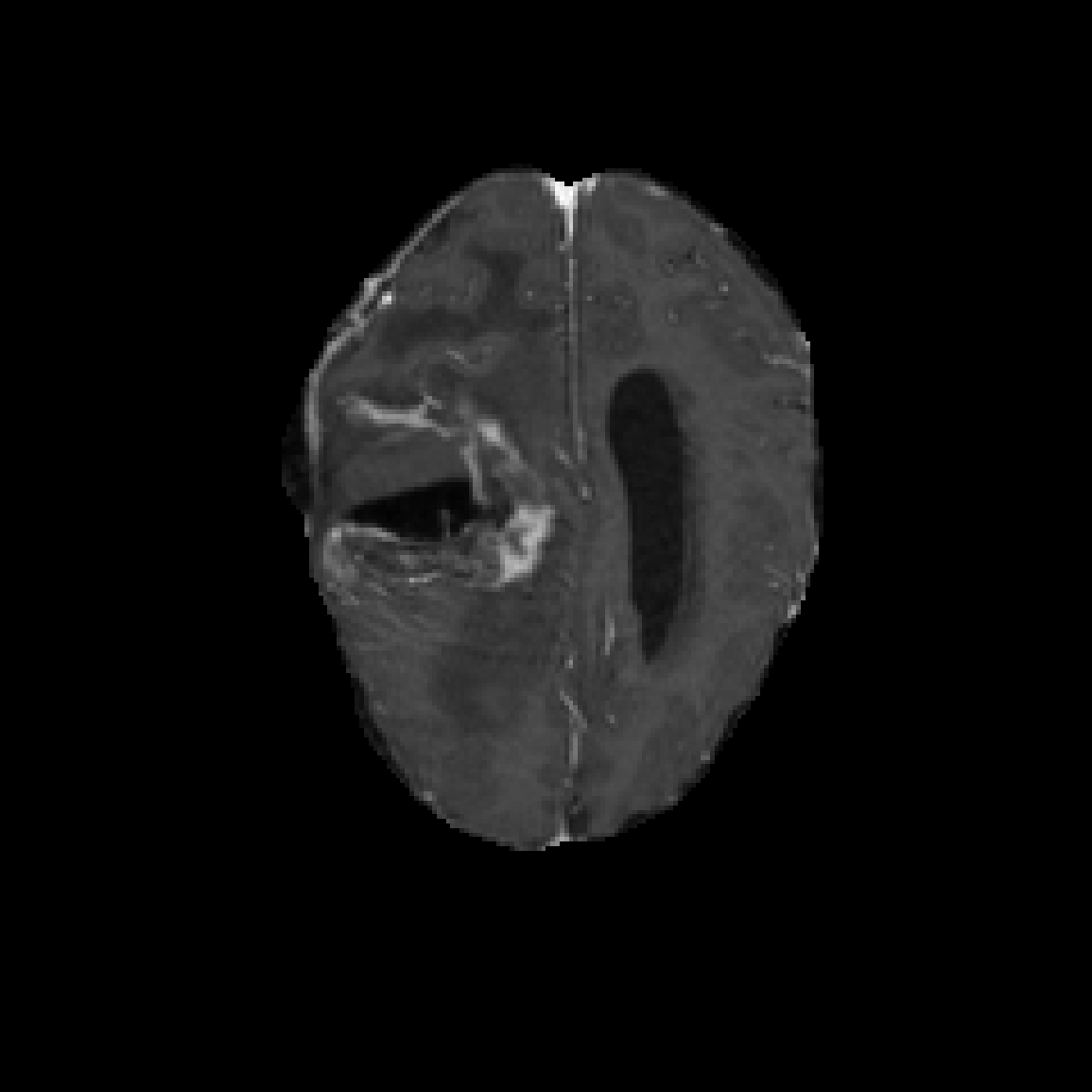} &
\includegraphics[width=0.18\textwidth]{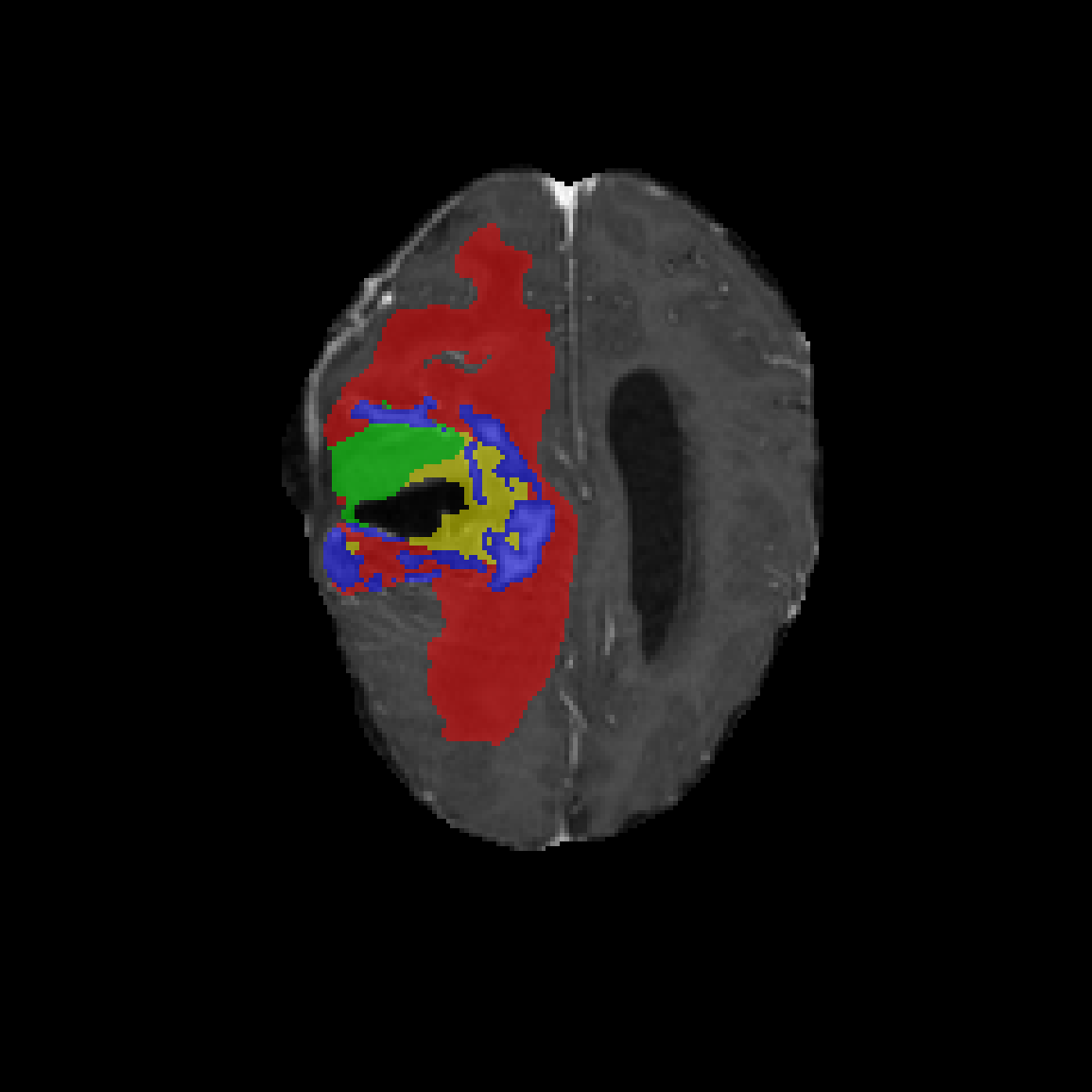} &
\includegraphics[width=0.18\textwidth]{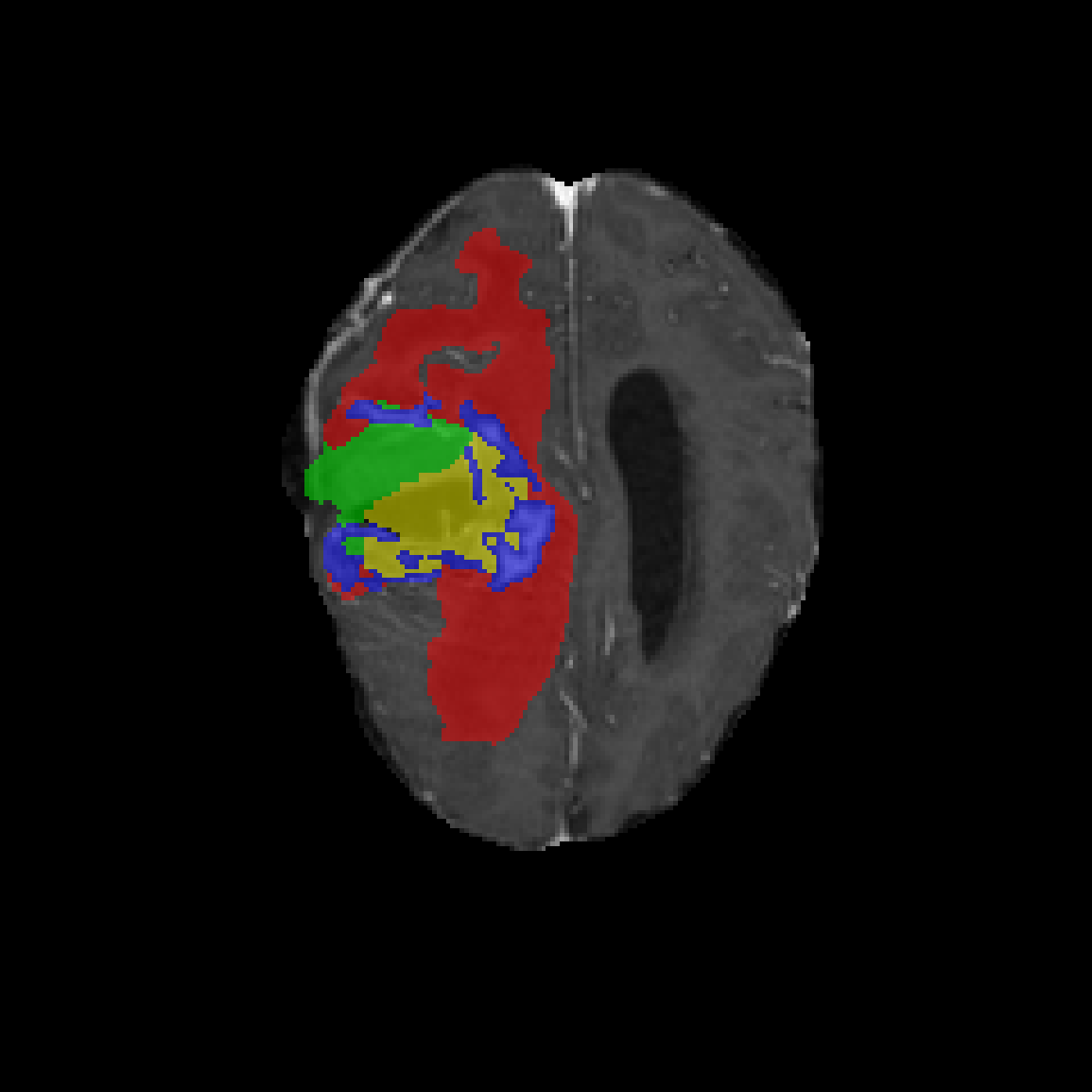} &
\includegraphics[width=0.18\textwidth]{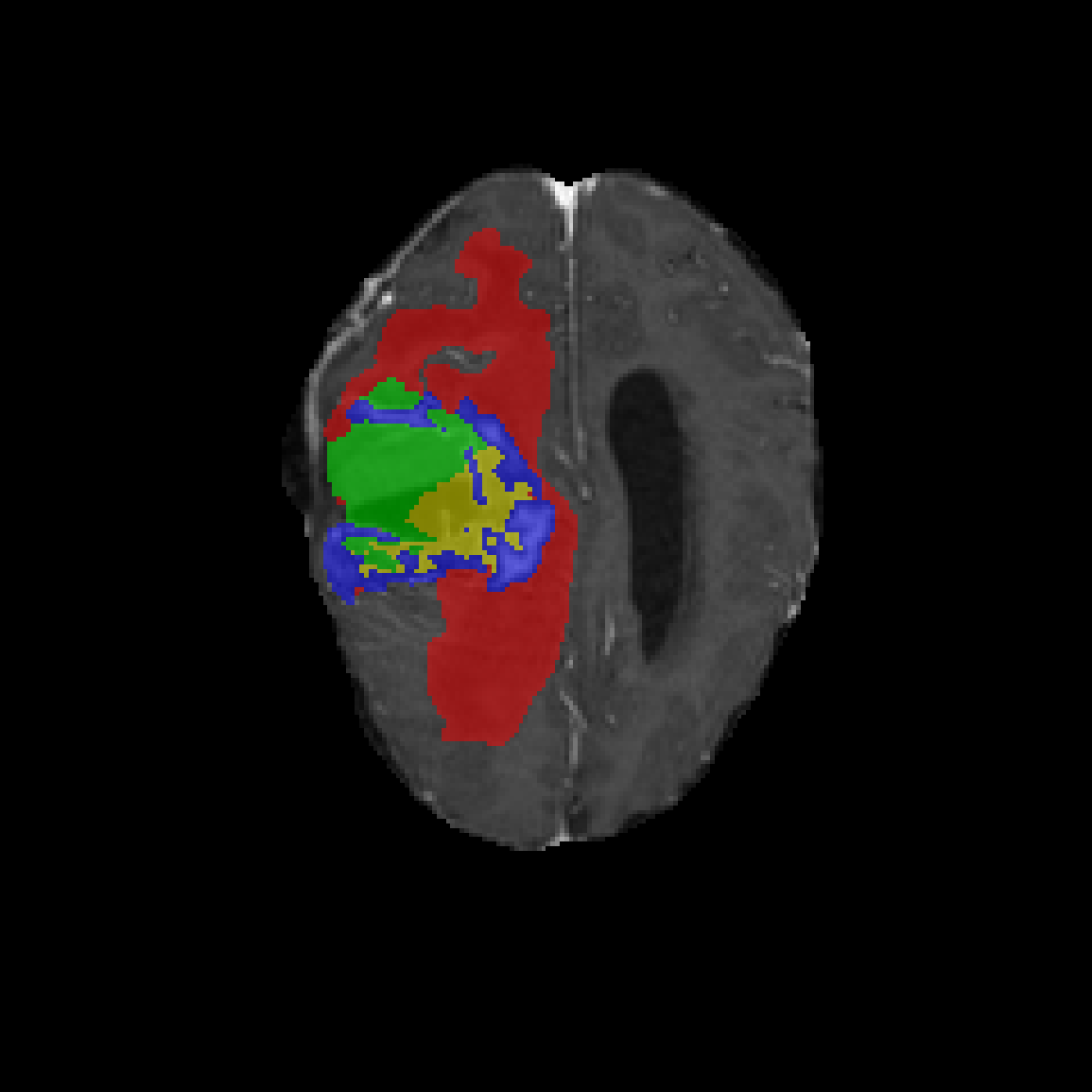} &
\includegraphics[width=0.18\textwidth]{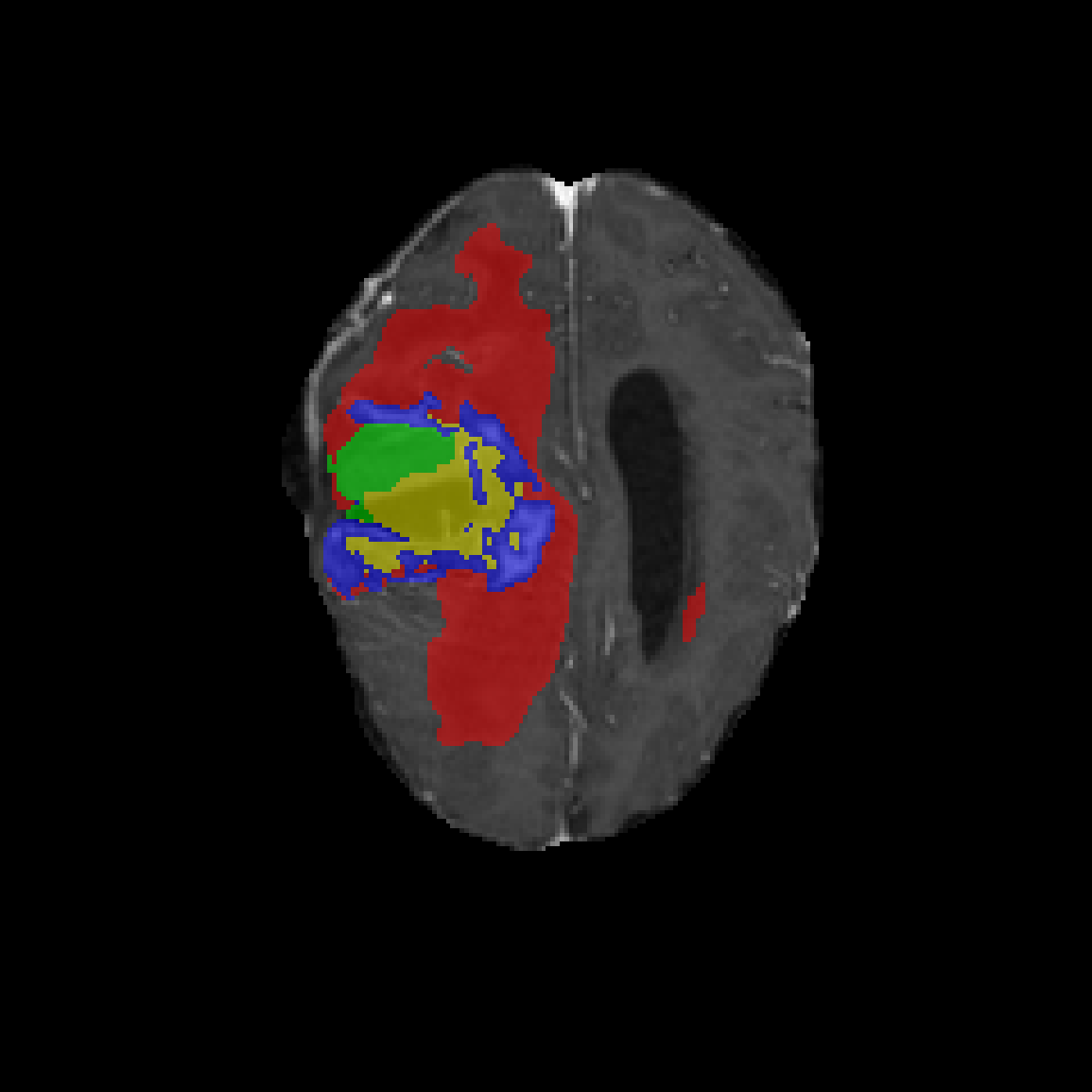} \\
\end{tabular}
\caption{Qualitative results. Columns (left to right): T1c axial slice, Ground
Truth (GT), \textit{Baseline}, \textit{Norm Variant}, and \textit{SACA Variant}.
Segmentation labels: ET (blue), TC (yellow), RC (green), and SNFH (red).
Case 1: WL Dice $> 0.95$; Case 2: WL Dice $\approx 0.80$; Case 3: morphology
featuring a large resection cavity.}
\label{fig:qualitative}
\vspace{-13pt}
\end{figure}

Figure~\ref{fig:qualitative}  shows segmentation overlays for three representative post-operative cases: a high-performing instance
(WL Dice $> 0.95$), a median case (WL Dice $\approx 0.80$), and a difficult case with a large resection cavity. In Case~1, all configurations capture the main tumor topology. Cases~2 and~3 show where the models struggle: around post-surgical white matter hyperintensity and complex resection-cavity anatomy, respectively.

\subsection{Discussion}
\textbf{Loss function stability and the GDL paradox.}
The instability of GDL under domain shift stems from its automatic volume-based
class weighting, which encodes scanner-specific tissue proportion statistics as
implicit priors. Under distributional shift, this mechanism guides the optimizer
with a corrupted signal that no longer reflects transferable anatomical structure.
Static class weights, as in the \textit{DiceWCE} objective, decouple the loss
from volume statistics entirely, providing a more stable optimization landscape
across heterogeneous acquisition protocols. This bias also explains the
degradation observed in the ensemble: systematic prediction errors do not cancel
upon probability averaging but instead corrupt the stronger model's posterior
estimates, making a biased model an actively harmful ensemble component.

\textbf{SACA, normalization, and their trade-offs.}
Post-operative MRI contains several competing signals: surgical debris can look like enhancing tumor on T1c, and post-surgical FLAIR hyperintensity can be mistaken for peritumoral edema. The SACA module addresses this by applying
shared class-conditioned queries to structured channel subspaces per spatial
token, so each tumor phenotype is recalibrated on its own rather than through a single shared bottleneck. Brain-masked percentile clipping plays a similar role on the input side: it anchors normalization to valid tissue and limits the influence of intensity outliers from blood products and resection cavities, while the contrastive loss pushes the features toward being hardware-invariant.
Both cause optimization complexity---subspace attention needs more signal to converge, and the contrastive loss competes with the segmentation gradient early in training---which is the likely reason their overlap scores on fine-grained sub-regions sit slightly below the \textit{Baseline}. For post-operative monitoring, this is an acceptable trade-off: missing residual active tumor costs more than a localized over-segmentation~\cite{ellingson2017modified}.

\textbf{Ensemble synergy.}
SwinUNETR and nnU-Net are complementary because their inductive biases differ: shifted-window self-attention captures long-range context, while convolutions are better at local boundary detail. Averaging their posteriors lets each cover the other's weak spots, which is why the \textit{SACA Variant} ensemble gives the best boundary fidelity: the most relevant metric to surgical planning.

\section{Conclusion}

We have shown that reliable post-operative glioma segmentation depends as much on the training objective and data representation as on the architecture. Our ablation study isolates three factors:  loss function stability, subspace-aware feature calibration, and intensity normalization, and shows a weakness of the Generalized Dice Loss under domain shift that matters for any multi-site pipeline. The SACA module and percentile-based normalization are lightweight additions that improve active tumor sensitivity and cross-site consistency, and the employed stable hybrid loss prevents the collapse that GDL suffers under distributional shift. 
The resulting ensemble is competitive on an independent external test set, a step toward longitudinal monitoring that could be clinically deployed. The main limitation of this study is its single-site training; future work will extend to multiple sites and validate prospectively on larger cohorts.

\section*{Acknowledgement}
This research is supported by the project \emph{``Romanian Hub for Artificial Intelligence -- HRIA'', Smart Growth, Digitization and Financial Instruments Program, 2021--2027, MySMIS no.\ 351416}.

\bibliographystyle{splncs04}
\bibliography{references}

\end{document}